\documentclass[iicol,pdflatex,sn-mathphys-num]{sn-jnl}

\usepackage{mathrsfs} 
\usepackage{graphicx}%
\usepackage{multirow}%
\usepackage{amsmath,amssymb,amsfonts}%
\usepackage{amsthm}%

\usepackage[title]{appendix}%
\usepackage{xcolor}%
\usepackage{textcomp}%
\usepackage{manyfoot}%
\usepackage{booktabs}%
\usepackage{algorithm}%
\usepackage{algorithmicx}%
\usepackage{algpseudocode}%
\usepackage{listings}%

\newcommand{\inp}[2]{\left\langle #1,\, #2\right\rangle}
\newcommand{\norm}[1]{\left\lVert #1\right\rVert}
\newcommand{\abs}[1]{\left\lvert #1\right\rvert}
\usepackage{mathtools}          
\allowdisplaybreaks             

\theoremstyle{thmstyleone}%
\newtheorem{theorem}{Theorem}
\newtheorem{proposition}[theorem]{Proposition}%

\theoremstyle{thmstyletwo}%

\theoremstyle{thmstylethree}%

\begin{document}

\title[Aticle Title]{Identifying Memorization of Diffusion Models through $p$-Laplace Analysis: Estimators, Bounds and Applications}


\author*[1,2]{\fnm{Jonathan} \sur{Brokman}}\email{jonathabrok@gmail.com}
\equalcont{These authors contributed equally to this work.}

\author[2]{\fnm{Itay} \sur{Gershon}}\email{itay.gershon@fujitsu.com}
\equalcont{These authors contributed equally to this work.}

\author[2]{\fnm{Amit} \sur{Giloni}}\email{amit.giloni@fujitsu.com}

\author[2]{\fnm{Omer} \sur{Hofman}}\email{omer.hofman@fujitsu.com}

\author[2]{\fnm{Roman} \sur{Vainshtein}}\email{roman.vainshtein@fujitsu.com}

\author[3]{\fnm{Kojima} \sur{Hisashi}}\email{roman.vainshtein@fujitsu.com}

\author[1]{\fnm{Guy} \sur{Gilboa}}\email{guy.gilboa@ee.technion.ac.il}

\affil*[1]{\orgdiv{Electrical and Computer Engineering}, \orgname{Technion}}

\affil[2]{\orgname{Fujitsu Research of Europe}}

\affil[3]{\orgname{Fujitsu Research}}


\abstract{Diffusion models, today's leading image generative models, estimate the score function, i.e. the gradient of the log probability of (perturbed) data samples, without direct access to the underlying probability distribution. This work investigates whether the estimated score function can be leveraged to compute higher-order differentials, namely the p-Laplace operators. We show that these operators can be employed to identify memorized training data. We propose a numerical p-Laplace approximation based on the learned score functions, showing its effectiveness in identifying key features of the probability landscape. \textcolor{black}{Furthermore, theoretical error-bounds to these estimators are proven and demonstrated numerically}. We analyze the structured case of Gaussian mixture models, and demonstrate that the results carry-over to text-conditioned image generative models (text-to-image), where memorization identification based on the p-Laplace operator is performed for the first time\textcolor{black}{, showing its advantage on 500 memorized prompts ($\sim$3000 generated images) in a post-generation regime, especially when the conditioning text is unavailable}.}




\maketitle

\section{Introduction}

Memorization in generative models occurs when the model produces outputs that closely replicate samples from its training data, rather than generating novel content. This issue is notably present even in contemporary popular diffusion models. This highlights important questions about generalization, and also raises privacy concerns in generative AI, as it can lead to the unintended leakage of sensitive training data~\cite{jahanian2021generative, somepalli2023diffusion}.

A key aspect of memorization lies in its relationship with the probability density of data points. Research findings often link memorization to ``bumps'' or ``delta regions'' in the learned data distribution. These bumps  may arise in sparsely populated areas of the training dataset due to limited data size. They also occur in inherently low-probability regions (of valid data samples), or result from unintended amplification of probability mass around replicated samples within the training set \cite{feldman2020does}, \cite{carlini2023extracting}, \cite{kadkhodaie2023generalization}.

The $p$-Laplace is a fundamental differential operator which can be used to quantify degrees of smoothness (also in high dimensions, as desired for our case). We outline below some of its essential properties. Recall the discrete Hilbert space $\mathbb{R}^d$ with standard discrete Euclidean inner product and norm definitions, i.e. $\inp{u}{v} =\sum_{i}u_{i}\cdot v_{i}$ and $\norm{u}^2 = \inp{u}{u}$, respectively. 
The $p$-Dirichlet energy for a function $u: \mathbb{R}^d \rightarrow \mathbb{R}$ is defined by,
\begin{equation}\label{eq:pDirichletEnergy}
    \begin{split}
    J_p(u):=&\frac{1}{p}\inp{\abs{\nabla u}^p}{1}=\frac{1}{p}\inp{\abs{\nabla u}^{p-2}\nabla u}{\nabla u}\\
    &=\frac{1}{p}\inp{-\textrm{div}\left(\abs{\nabla u}^{p-2}\nabla u\right)}{ u}\\
    =&\frac{1}{p}\inp{-\Delta_p u}{u},
    \end{split}
\end{equation}
where  $\Delta_p u$ is the $p$-Laplace operator defined as 
\begin{equation}\label{eq:pLaplacian}
    \Delta_p u := \nabla \cdot \left(\abs{\nabla u}^{p-2}\nabla u\right),
\end{equation}
$\nabla \cdot$ and $\nabla$ are the discrete divergence and gradient operators respectively, and $\left| \nabla u \right|^{p-2} = \left[ \left( \frac{\partial u}{\partial x(1)} \right)^2 + \cdots + \left( \frac{\partial u}{\partial x(d)} \right)^2 \right]^{\frac{p-2}{2}}
$ . We note that $-\Delta_p$ is a positive semi-definite operator and that $J_p(u)\ge 0,\, \forall u$. 
The $p$-Laplace is $p-1$ homogeneous i.e.
\begin{equation}\label{eq:homogeneity}
    \Delta_p(a\cdot u)=a\abs{a}^{p-2}\cdot\Delta_p(u),\quad \forall a\in\mathbb{R}.
\end{equation}
We focus on function analysis via the $p$-Laplace and its ability to reliably characterize important phenomena. 

We investigate the ability to use the $p$-Laplace to characterize memorized samples in diffusion models. In this setting, the probability distribution is that of natural image data, which is unknown. To address this challenge we propose to capture the ``probability bump'' around memorized samples using $p$-Laplace approximations, a technique not yet explored. This leads us to several fundamental questions:
\begin{itemize}
    \item How can we estimate the $p$-Laplace operator in the settings of diffusion models?
    \item \textcolor{black}{Can we bound the estimation error for reliability?}
    \item What estimation techniques and values of \( p \) are best suited for practical applications? 
    \item Can the $p$-Laplace operator effectively characterize points of memorization?
\end{itemize}

Our findings outline both recommended and less desired approaches for this task, thereby advancing the research on the geometry of the implicitly learned probability function in diffusion models. We provide a novel mathematical perspective that  supports the concept of a ``bumpy probability'' around memorized samples\textcolor{black}{, and verify its applicability in identifying this phenomenon on 500 memorized prompts ($\sim$3000 generated images)}. \footnote{Official implementation:

\url{https://github.com/JonathanBrok/Identifying-Memorization-of-Diffusion-Models-through-p-Laplace-Analysis/} }

\section{Previous Work}
Memorization is a well-researched phenomenon. The work of  \cite{feldman2020does} provided theory for long-tailed distributions, claiming that memorizing rare samples from low-probability regions enhances generalization. In \cite{feldman2020neural} they further showed that neural networks allocate probability mass to these sparse instances. In the context of diffusion models, \cite{kadkhodaie2023generalization} observed that limited data for diffusion model training causes memorization, and analyzed the generalization that occurs as the data increases in size. An alternative research branch, such as \cite{carlini2023extracting} and \cite{somepalli2023diffusion}, focused on  data replication in diffusion model training sets and its effect of memorization amplification - exposing the possibility of unintended data leakage, raising privacy and copyrights concerns. \textcolor{black}{Recently, \cite{wen2024detecting} proposed a method to identify\footnote{Identifying memorization is often termed \textit{detecting} memorization in the literature} and mitigate this phenomenon through score analysis. Contemporaneous works: \cite{jeon2025understanding} studied a Hessian-based approach for this task. In \cite{brokman2025tracking} we extend our geometric $p$-Laplace analysis to conditional probabilities, assuming access to the conditioning prompt. Applying the same framework to the conditional log-density, they report enhanced separability between memorized and non-memorized samples throughout the diffusion-model generative process.}

Effective uses of the $p$-Laplace in image processing is demonstrated in \cite{baravdish2015backward,chen2010image,chen2006variable,huang2016level,liu2016renormalized} for $1 \leq p \leq 2$ with $p$ dependent on the smoothness of the image.
Properties of the $p$-Laplace gradient flow in the context of image filtering are discussed in \cite{kuijper2013image,kuijper2007p,wei2012p,fazeny2023hypergraph}. For $p \to 1$, the energy approaches total-variation and the gradient flow approximates the total variation (TV) flow \cite{andreu2001minimizing}.
Properties of the graph $p$-Laplace were analyzed in \cite{elmoataz2015p,hafiene2018nonlocal,slepcev2019analysis,brokman2021nonlinear,weihs2024discrete,brokman2024spectral}, showing several data-processing applications, including for message passing in pre-trained image-language models \cite{wu2024p}.
Analysis of the operator as $p \to \infty$ was performed in \cite{bungert2023inhomogeneous,deidda2024graph}. Nonlinear $p$-Laplace eigenvalue problems and the $p$-Laplace spectra were investigated in \cite{cohen2020introducing,bozorgnia2024infinity}. 

Nonlinear operators and PDEs (Partial Differential Equations) have been studied in contexts related to generative modeling and diffusion processes. For instance, \cite{albergo2019flow} introduced flow-based generative models for lattice field theory, using PDE frameworks. \cite{lim2023score} extended score-based diffusion models to function spaces, showcasing the flexibility of diffusion models in handling complex data structures and function representations. Recently, \cite{anonymous2024manifold} tackled generated versus real image detection using the cross-similarity between predicted noise and signal, which they re-interpreted as a combined criterion of curvature (1-Laplace), gradient, and bias of the noise predictions.

To the best of our knowledge,  we are the first to estimate the $p$-Laplace with diffusion models and to use it to characterize the learned probability distribution. Namely, we are the first to leverage $p$-Laplace for memorization analysis.



\noindent\textbf{Relationship to the Conference Version.}\label{sec:related_work_relationship}
\textcolor{purple}{
This article is an extended version of our conference paper presented at SSVM'25~\cite{brokman2025identifying}. The SSVM version introduced the use of the $p$-Laplace of the learned log-density to expose memorization “bumps”, proposed estimators from learned scores, and provided evidence on the advantage of the $p{=}1$ boundary formulation
. The present journal version adds (i) \emph{theoretical error bounds} of the estimators
, with constants depending on $p$ and norm bounds; (ii) clarifying the small-$\alpha$ post-generation regime
; and (iii) a \emph{substantially expanded} empirical study: A large-scale text-to-image evaluation on 500 memorized prompts ($\sim$3{,}000 images) including comparisons to the recent~\cite{wen2024detecting}. 
. We also provide additional qualitative results and full proofs of the new bounds. 
} 

\section{Method}

This section is structured as follows: Sec. \ref{sec:method_hyp} states the core hypothesis motivating our approach. Sec. \ref{sec:diffusion_settings} describes the diffusion model formulation and the regimes in which we analyze scores. Sec. \ref{sec:method} introduces our $p$-Laplace estimator and its formulations. \textcolor{black}{Finally, Sec. \ref{subsec:error_bound} provides error bounds on our estimators and discusses their reliability (proofs in the Appendix \ref{sec:err_bound_proof})}.




\subsection{Hypothesis: Measuring Memorization with $p$-Laplace}\label{sec:method_hyp}
Building upon prior observations that relate memorized samples to ``bumps'' in the training data distribution, we hypothesize that these memorized samples will manifest as local maxima in the learned (log) probability distribution. The \(p\)-Laplace, which quantifies the outward flux of \(|\nabla u|^{p-2} \nabla u\) around a point (as elaborated below in Sec. \ref{sec:method}), serves as a tool for this analysis. Typically, around local maxima, the gradient vectors point inward, indicating a negative flux. Therefore, we expect that memorized points will exhibit lower (more negative, with higher absolute value) values of the \(p\)-Laplace for their learned probabilities.

\subsection{Diffusion Model Settings}\label{sec:diffusion_settings}
Given data in $\mathbb{R}^d$ with unknown density, diffusion models learn the \emph{score function}
$s_t(x) := \nabla \log p_t(x)$ of the noise-perturbed density $p_t$, where $t$ indexes the noise level and $x$ is the perturbed data sample.

\subsection{Score learning via noise prediction.}
For the Gaussian corruption
\begin{equation}\label{eq:alpha_perturb}
x_t=\sqrt{1-\alpha_t}\,x_0+\sqrt{\alpha_t}\,\epsilon,\quad \epsilon\sim\mathcal N(0,I),   
\end{equation}
we have\footnote{$\mathcal{N}(a;b,c)$ denotes the probability density function of a Gaussian RV $a$ with mean and covariance $b,c$ - this is standard notation in the diffusion models literature}
\begin{equation}\label{eq:probdens}
q(x_t\mid x_0)=\mathcal N\!\big(x_t;\ \sqrt{1-\alpha_t}\,x_0,\ \alpha_t I\big).
\end{equation}
The MMSE predictor of the noise satisfies
\[
\mathbb E[\epsilon\mid x_t] \;=\; -\,\sqrt{\alpha_t}\,\nabla \log p_t(x_t),
\]
so predicting $\epsilon$ is equivalent (up to the known factor $\sqrt{\alpha_t}$) to estimating the score $s_t(x_t)$ \cite{miyasawa1961empirical,song2019generative,nichol2021improved}.

\paragraph{Forward and reverse diffusions (VP SDE).}
In the small-step limit, the forward process can be shown via Taylor to be
\[
d x_t \;=\; -\tfrac{1}{2}\,\beta(t)\,x_t\,dt \;+\; \sqrt{\beta(t)}\,dW_t,
\]
where \(W_t\) is a Wiener process (Brownian motion), and with $\beta(t)\ge 0$ and
\[
\alpha(t)=\int_0^t \beta(s)\,ds.
\]
The associated Fokker--Planck equation governs $p_t(x)$, and provides the reverse-time SDE 
\[
d x_t \;=\; \Big[-\tfrac{1}{2}\beta(t)\,x_t \;-\; \beta(t)\,\nabla \log p_t(x_t)\Big]\,dt
\;+\; \sqrt{\beta(t)}\,dW_t.
\]
This  is reverse in the sense that if $\tilde p_t(x)$ denotes the time-varying density of the reverse process, then $\tilde p_t(x)=p_{T-t}(x)$. Note how the reverse process traces samples from pure Gaussian back to samples from the clean data density.

\textcolor{black}{
\paragraph{Small–$\alpha$ regime.} Let us fix $t$ alongside its corresponding noise scale and density, and switch notation to $\alpha \leftarrow \alpha_t, p \leftarrow p_t$. Then assuming we have access to a score $s(x)=\nabla\log p(x)$ (or an accurate estimator), we can simulate gradient ascent on $\log p$ with this mild Gaussian perturbation. This has two benefits we exploit: (i) it preserves \emph{regularity} by effectively sampling from a slightly smooth version of the density (more on that below); (ii) it remains \emph{geometrically faithful} to local structures of $p$ (e.g., attraction basins around local maxima). Thus we use the “last–step denoising” regime in practice - matching the final diffusion step where the noise level is smallest. This alignment lets us study memorization in a setting where the sampler perturbs minimally while still ensuring a well-defined derivatives, and is essentially a \textit{post-generation} regime, meaning that we inspect a generated sample after it was generated (as opposed to identifying memorization mid-generation).}

Importantly, this entire generative process and its analysis often takes place in latent space, as introduced by \cite{rombach2022high}. 
In their work, they proposed to run diffusion in a learned latent space rather than in pixel space. First, a Variational Autoencoder (VAE) maps data 
\[
x \in \mathbb{R}^d 
\quad\longmapsto\quad 
z = E(x) \in \mathbb{R}^m,
\]
where \(E\) is the encoder. The forward noising in latent space is then
\begin{equation}\label{eq:latentForward}
z_{t} = \sqrt{1 - \alpha_t}\,z_{0} + \sqrt{\alpha_t}\,\epsilon, 
\quad \epsilon \sim \mathcal{N}(0,I),
\end{equation}
and the model is trained to invert \eqref{eq:latentForward} via noise prediction. After denoising, the final latent sample \(z_0\) is decoded to pixel space:
\begin{equation}\label{eq:latentDecoding}
\widehat{x}_0 = D(z_0),
\end{equation}
where \(D\) is the VAE decoder.  

Working in latent space lowers compute costs and speeds up training, compared to direct diffusion on high-resolution images. Artifacts may appear due to imperfect VAE reconstruction, yet our $p$-Laplace analysis, introduced below, applies if we treat the learned latent distribution as $p(z)$ and measure memorized ``bumps'' within that domain.


\subsection{Method: $p$-Laplace Estimation}\label{sec:method}

Consider  \(u(x) = \log p(x)\), and consequently \(s(x) = \nabla u(x)\). Denote \(\hat{s}(x)\) as the approximation of $s$ given by the diffusion model - i.e. the relationships are:
\begin{equation}\label{eq:gradient_approx}
    \nabla u(x) = s(x) \approx \hat{s}(x).
\end{equation}
By Eq. \eqref{eq:gradient_approx}, the $p$-Laplace \(\Delta_p u\) can be written as:
\begin{equation}
    \Delta_p u(x) = \nabla \cdot \bigl( |s(x)|^{p-2} s(x) \bigr).
    \label{eq:p_laplacian}
\end{equation}
\paragraph{Volume and boundary integral formulations:}
To approximate \(\Delta_p u(x_0)\) locally, we consider a $d$-dimensional ball \(B_R(x_0)\) of radius \(R\). The average $p$-Laplace in the ball is:
\begin{align*}
\overline{\Delta_p u}(x_0) &= \frac{1}{|B_R(x_0)|}\int_{B_R(x_0)} \Delta_p u(x)\, dx\\
&= \frac{1}{|B_R(x_0)|}\int_{B_R(x_0)} \nabla \cdot \bigl( |s(x)|^{p-2} s(x) \bigr)\, dx,
\end{align*}
where \(|B_R(x_0)|\) is the volume of the ball. Alternatively, by the divergence theorem,
\[
\overline{\Delta_p u}(x_0) = \frac{1}{|B_R(x_0)|}\int_{\partial B_R(x_0)} |s(x)|^{p-2}s(x)\cdot n \, ds.
\]
with \(n\) the outward unit normal on the boundary \(\partial B_R(x_0)\).
\paragraph{d-Dimensional ball and sphere measures:}
The volume of a $d$-dimensional ball of radius \(R\) is:
\[
|B_R(x_0)| = \frac{\pi^{d/2}}{\Gamma(\tfrac{d}{2}+1)} R^d,
\]
where $\Gamma$ denotes the gamma function. The surface measure of the d-sphere \(S^{d-1}\) of radius \(R\) is:
\[
|\partial B_R(x_0)| = \frac{2\pi^{d/2}}{\Gamma(d/2)} R^{d-1}.
\]

\paragraph{Discrete Monte Carlo approximation:}
Since we do not have closed-form integrals, we approximate these integrals numerically:

\noindent1. {\bf Volume Integral Approximation:}  
   Uniformly sample \(N\) points \(\{x_i\}\) inside \(B_R(x_0)\). Compute \(\Delta_p u(x_i)\) at each sampled point:
   \begin{equation}\label{eq:volume_formulation}
   \overline{\Delta_p u}(x_0) \approx \frac{1}{N}\sum_{i=1}^N\nabla \cdot \bigl( |s(x)|^{p-2} s(x)).
   \end{equation}

\noindent2. {\bf Boundary Integral Approximation:}  
   Uniformly sample \(N\) points \(\{y_i\}\) on the sphere \(\partial B_R(x_0)\). Evaluate \(|s(y_i)|^{p-2}s(y_i)\cdot n_i\) and sum:
   \begin{equation}\label{eq:boundary_formulation}
   \overline{\Delta_p u}(x_0) \approx \frac{|\partial B_R(x_0)|}{|B_R(x_0)|}\left(\frac{1}{N}\sum_{i=1}^N |s(y_i)|^{p-2}s(y_i)\cdot n_i \right).
   \end{equation}
The $\frac{|\partial B_R(x_0)|}{|B_R(x_0)|}$ factor was introduced since without it the Monte Carlo converges to a mean over the sphere $\partial B_R(x_0)$. This factor ensures that we do not normalize by the sphere's surface, and correctly normalize by the volume.

\paragraph{Regularity for Applying the Divergence Theorem.}
The perturbations in Eq.\ \eqref{eq:alpha_perturb} induce the probability density
\begin{align*}\label{eq:gaussian_convolution_alpha}
p_{\alpha}(x) &= (\sqrt{1-\alpha}p_0 \;*\; \mathcal{N}(x;\sqrt{1-\alpha}x_0, \alpha I)\\
&=\sqrt{1-\alpha}\int_{\mathbb{R}^d}
p_0(y)\;G_{\alpha}(x - y)\,\mathrm{d}y,
\end{align*}
where \(p_0\) is the underlying (unperturbed)  and $\mathcal{N}$ denotes the Gaussian probability density function similarly to Eq. \eqref{eq:probdens}.  As is well known, convolving any locally integrable density (and assuming \(p_0\) is such) with a Gaussian kernel yields a \(C^{\infty}\) function on \(\mathbb{R}^d\) -  (see for instance "properties of mollifications" in \cite{adams2003sobolev}).  Hence, the resulting \(p_{\alpha}(x)\) is smooth for all \(\alpha>0\), and strictly positive where \(p_0\) is non-zero.  Therefore, the log-likelihood inherits sufficient regularity for the divergence theorem to apply in our derivations.  

\paragraph{Using score from diffusion model:}
In practice, \(s(x)\) may be unknown. If we have a trained diffusion model that estimates the score function, we plug in its approximation \(\hat{s}(x)\) instead of \(s(x)\). Note that this introduces an additional degree of approximation to Eqs. \eqref{eq:volume_formulation}, \eqref{eq:boundary_formulation}. Moreover, using $p_\alpha$ means we use a smoothed version of the original density - thus set a small $\alpha$. As a rule of thumb we set it to the smallest value that the diffusion model is designed to deal with - which is the $\alpha$ of the last denoising step. 

By comparing $p$-Laplace estimates computed from the true score (if available) and the learned score, we assess the fidelity of the diffusion model $p$-Laplace approximation. We also compare between the volume-based and boundary-based formulations, and, finally, experiment with $p$-Laplace upon memorization of the diffusion model.



\subsection{Error Bounds}\label{subsec:error_bound}

\textcolor{black}{Diffusion models offer only an approximation of the score function; therefore, it is essential to measure the error of our estimators based on this approximation. To achieve this, we derive and empirically validate error bounds on our estimators.} 

\textcolor{black}{
Let us fix $p \in[1, \infty], R>0$, and $x_0\in \mathbb{R}^d$.
We wish to bound the error of the $p$-Laplace average flux approximation on $\partial B_R(x_0)$. Namely, we wish to bound \begin{equation}
    \abs{\overline{\Delta_p s}(x_0) - \overline{\Delta_p \hat{s}}(x_0)}.
\end{equation}
\begin{proposition}
Assume there exist $\delta>0$ and $0<m\leq M$ s.t. $\forall y \in \partial B_R(x_0)$, the following hold:
\begin{enumerate}[(a)]
    \item $\norm{s(y)-\hat{s}(y)}<\delta$
    \item $m < \norm{s(y)}, \norm{\hat{s}(y)} \leq M$
    \item $\forall t\in[0,1],$ $\norm{t\cdot s(y)+(1-t)\cdot\hat{s}(y)} \geq m$
\end{enumerate}
Then $\abs{\overline{\Delta_p s}(x_0) - \overline{\Delta_p \hat{s}}(x_0)} \leq C_p$, where
\begin{equation}
    C_p =
 \begin{dcases} 
      \frac{\abs{\partial B_R(x_0)}}{\abs{B_R(x_0)}}\cdot\delta M^{p-2}(p-1)  & p\geq 2 \\
      \frac{\abs{\partial B_R(x_0)}}{\abs{B_R(x_0)}}\cdot\delta m^{p-2}(3-p) & p<2 
   \end{dcases}
\end{equation}
\end{proposition}
\noindent Proof can be found in the appendix.}

\section{Experiments}
In this section, we evaluate the proposed numerical approximations and their error bounds and test the efficacy of \(p\)-Laplace in measuring memorization. First, following \cite{song2019generative}, we conduct experiments using a two-dimensional Gaussian mixture model (GMM): We assess our numerical approximations, and explore the memorization phenomenon, identifying the best choice of approximation and order \(p\) of the \(p\)-Laplace for capturing memorized instances. \textcolor{black}{We then proceed in the same GMM regime to numerically demonstrate the effectiveness of our error bounds.}

\textcolor{black}{
The final experiment extends these methods to a full-scale diffusion model, applying the insights gained from the GMM experiments. We demonstrate that our findings generalize effectively to the unknown, complex, high-dimensional probability landscape of natural images.}

\subsection{GMM Setup}
In this setting we have true analytic expressions for the gradient and divergence, which enable us obtain error statistics, considering the underlying scalar function $u(\mathbf{x})$ to be the log-probability of the GMM. For all GMM experiments, we use 3 mixture components, each with covariance $\sigma^2 I$ and equal mixture weights, and $100$ samples per integral estimate (same number for both boundary and volume formulations). Aggregated statistics (when applicable) are across $100$ independent runs per neighborhood-circle and the $p$ parameter. Radius of integration spheres: $R=1.0$. Finite difference step for numerical derivatives is set to $\delta=10^{-3}$.  

\paragraph{GMM diffusion model.} We train a diffusion model on the GMM data with $1000$ samples. As architecture we choose a MLP with one hidden layer (128 neurons) and sinusoidal time embeddings discretized to $T=100$ timesteps. Training is done for 500 epochs of "vanilla" SGD with learning rate of $1e^{-3}$.





\begin{figure*}
\includegraphics[width=\textwidth]{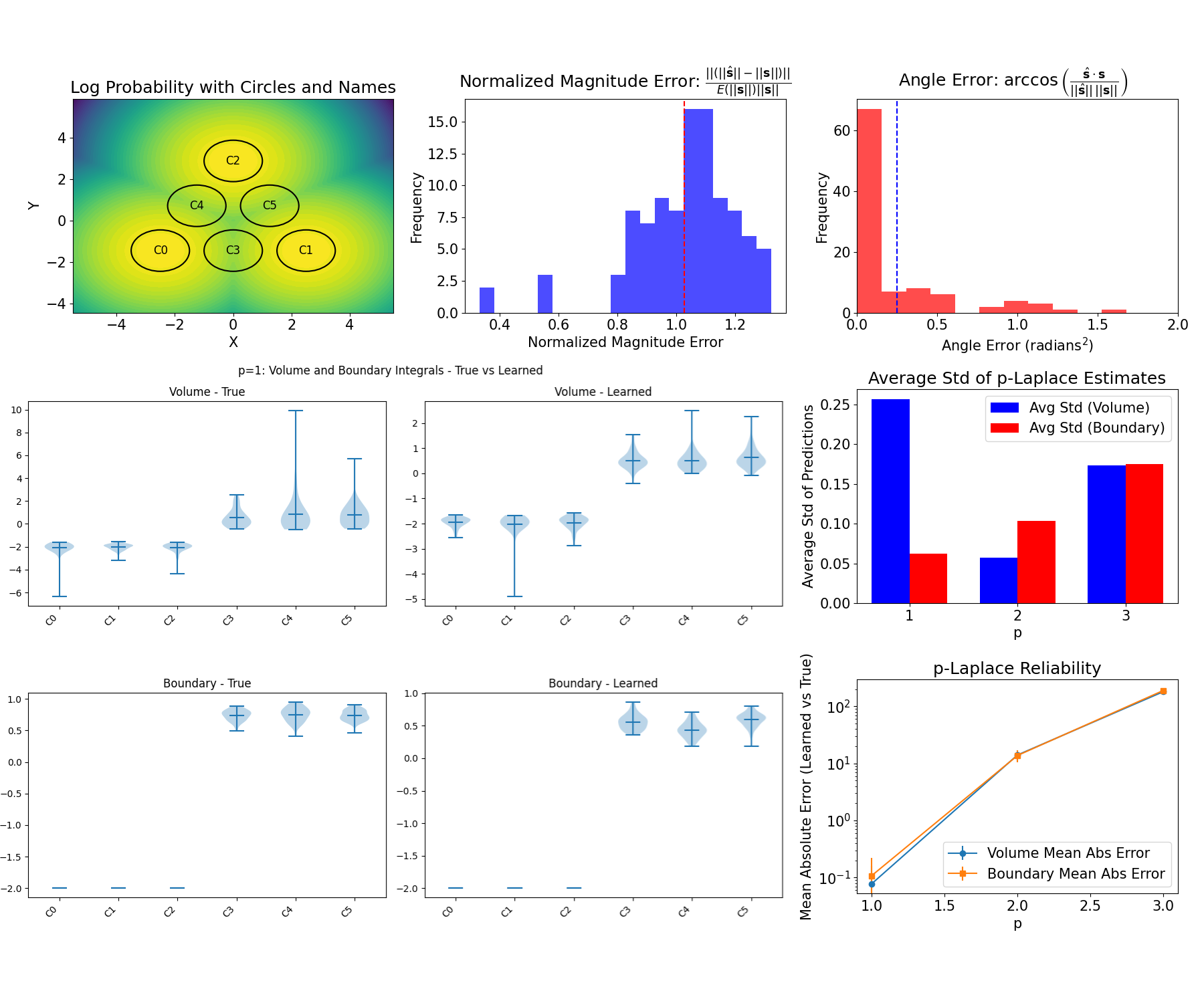}
\caption{GMM error analysis (take-away: $1$-Laplace boundary formulation is the most reliable). Here we test the fidelity of our approach, to the values of the accessible true probability and its $p$-Laplace. We test learned, and oracle fields, using volume and boundary formulations. \textbf{Top left:} 6 neighborhoods are tested - 3 maxima and 3 non-maxima points. \textbf{Top middle-right:} $\hat{s}$ achieves low error rate in direction estimation, however it is not the case for magnitudes, with relative error of $\sim \times 1.2$ receiving non-negligible frequency. \textbf{Bottom-right:} $1$-Laplace is approximated significantly better than in $p=2,3$ , due to invariance to errors in magnitude. \textbf{Middle-right:} The volume integral formulation has very high variance rendering it less reliable than the boundary formulation, especially for $p=1$. \textbf{Middle-bottom, left-middle:}  we provide further per-neighborhood violin plots for $p=1$ showing that, despite the high variance in volume formulation, maxima neighborhoods are still distinguishable from non-maxima.} \label{fig:gmm_err}
\end{figure*}

\subsection{Experiment 1: GMM Field Approximations}
In this experiment, we assess the correctness of our approach, where the error is estimated using access to true quantities: The true log probability density \(u(x)\)  and hence its $p$-Laplace \(\Delta_p u(x)\). This enables precise analysis and visualization - see Fig. \ref{fig:gmm_err}.

We compare various $p$-Laplace operators, namely $p=1,2,3$,  computed from: 
\begin{itemize}
    \item The \textbf{oracle} gradient field $s$ (direct differentiation of $u$).
    \item The \textbf{learned} gradient field $\hat{s}$, obtained by training a diffusion model on the GMM data.
\end{itemize}
We test  both boundary and volume integral formulations - Eqs. \eqref{eq:volume_formulation}, \eqref{eq:boundary_formulation} of Sec. \ref{sec:method} respectively.

Fig. \ref{fig:gmm_err} incorporates a thorough multi-faceted analysis, with a simple take-away: If we are interested in a reliable and robust $p$-Laplace, the best choice is the $1$-Laplace using boundary formulation for its approximation. We further reveal an important insight: In our experiments, $\hat{s}$ approximates the direction of $s$ far better than its magnitude. $1$-Laplace only uses normalized gradients, canceling the imprecise magnitude estimations - which indeed leads to its superior performance.



\begin{figure*}[htb]
\includegraphics[width=\textwidth]{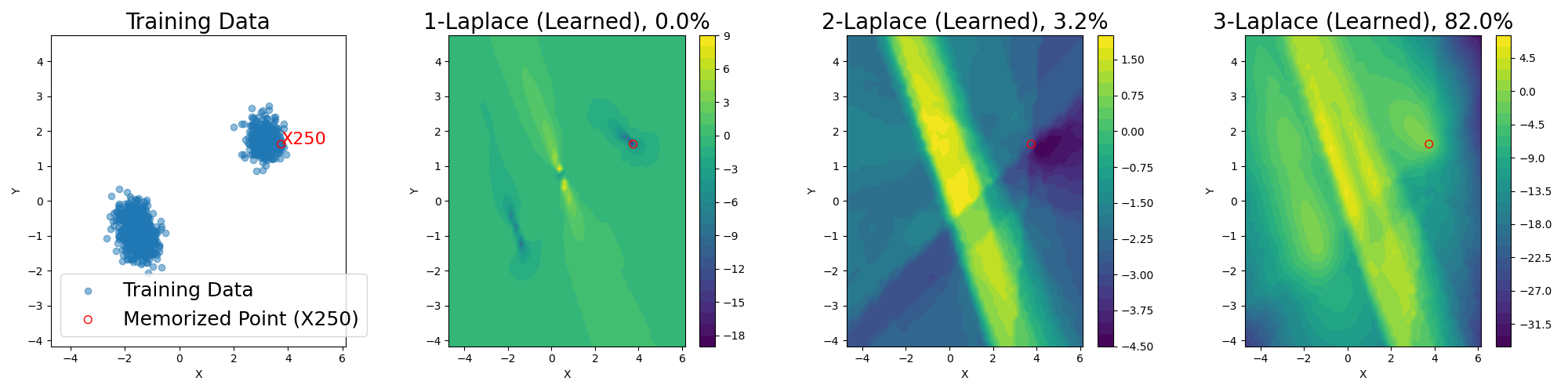}
\includegraphics[width=\textwidth]{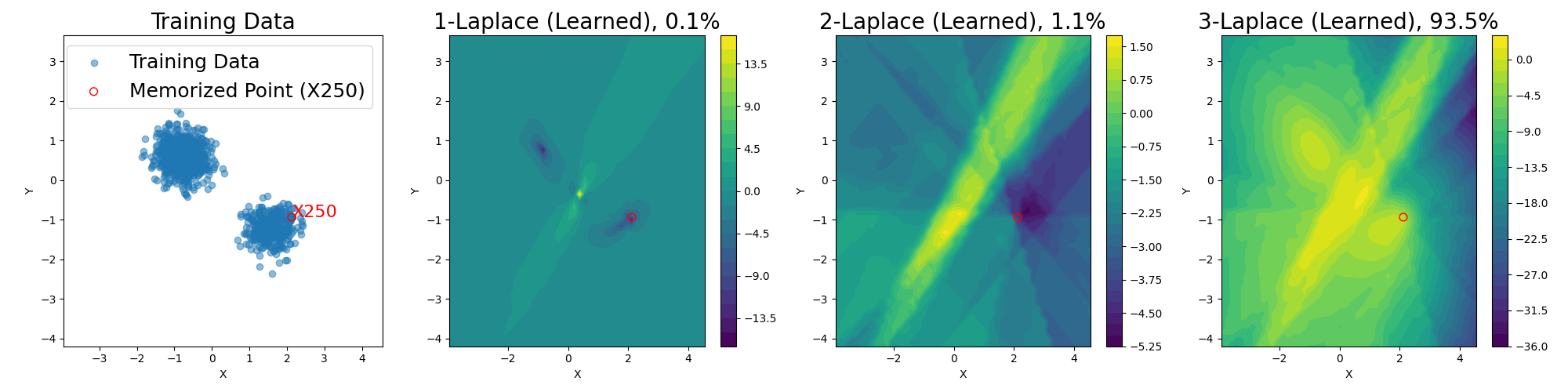}
\includegraphics[width=\textwidth]{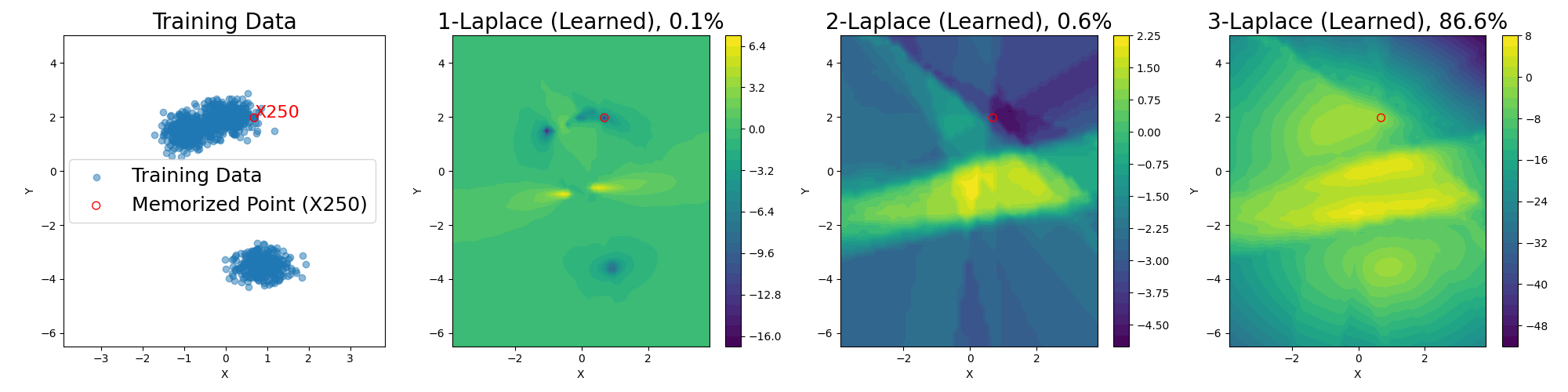}
\caption{We are interested in the ability of the $p$-Laplace of the implicitly learned log probability to reliably reflect memorization, and provide distinguishable values at memorized points. In each row, the scatter-plot shows the GMM training set with randomly drawn peak locations, including one red sample that was replicated 250 times to induce  memorization. The colormaps portray the $p$-Laplace, where we see again how the values of the $1$-Laplace pin-point better the memorized sample, assigning it the lowest percentile (percentages in title) compared to the other $p=2,3$-Laplace.} \label{fig:gmm_memorization}
\end{figure*}

\begin{figure*}[htb]
\centering
\includegraphics[width=\textwidth]{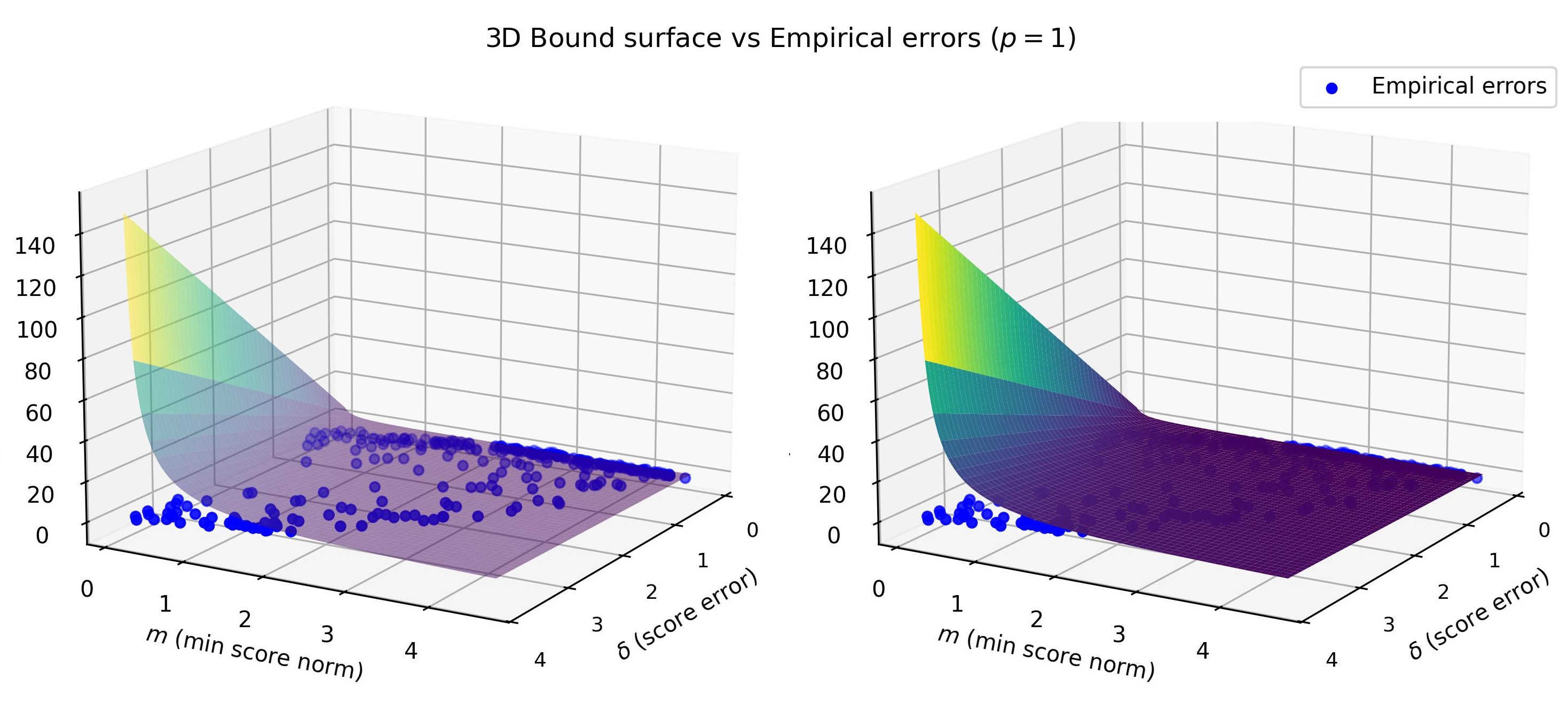}
\caption{\textcolor{black}{The theoretical error bound surface together with empirical errors showm in dots for the 2D GMM setting. The surface shows the analytic bound as a function of the score error $\delta$ and minimum score norm $m$, while blue points represent the empirical errors. The two subplots are the same with different opacity - to visually show that the error bound lies completely above the empirical errors - even in cases where the bound is tight.}}
\label{fig:error_bound_gmm}
\end{figure*}

\subsection{Experiment 2: Memorization Detection in GMM Setups}
Here we explore whether our approach for the $p$-Laplace approximations can distinguish memorized points from non-memorized ones. Since memorization is a synthetic phenomenon of the model, it is not captured by the true distribution - hence we will not experiment with the true distribution here.

We introduce a controlled memorization scenario within the GMM framework. By intentionally replicating a single training sample multiple times (adding 250 duplicates of one point to a 1000-sample training set), we induce a synthetic “spike” in the model’s learned distribution (Fig. \ref{fig:gmm_memorization}). Such replications were indeed previously observed to cause memorization in diffusion models, e.g. \cite{carlini2023extracting}.

The true distribution is naturally unaffected by the data sampling process, and consequently the true  $p$-Laplace is not expected to exhibit any anomaly at the memorized point (since the true distribution remains unchanged), whereas the \emph{learned} $p$-Laplace potentially reveals a distinct signature at the memorized location.

To quantify detectability, we:
\begin{itemize}
    \item Train the diffusion model on 1000 GMM samples plus 250 replicates of a randomly chosen point, simulating memorization.
    \item Compute the learned $p$-Laplace across a grid points
    \item Assess the percentile rank of the memorized point’s $p$-Laplace value among all grid points.
\end{itemize}

The results of this analysis, comparing $p$-Laplace of $p=1,2,3$, reveals that $p=1$ best highlights the memorized samples as outliers, as these samples occupy very low percentile values, see Fig. \ref{fig:gmm_memorization}.

\begin{figure*}[htb]
\centering
\includegraphics[width=\textwidth]{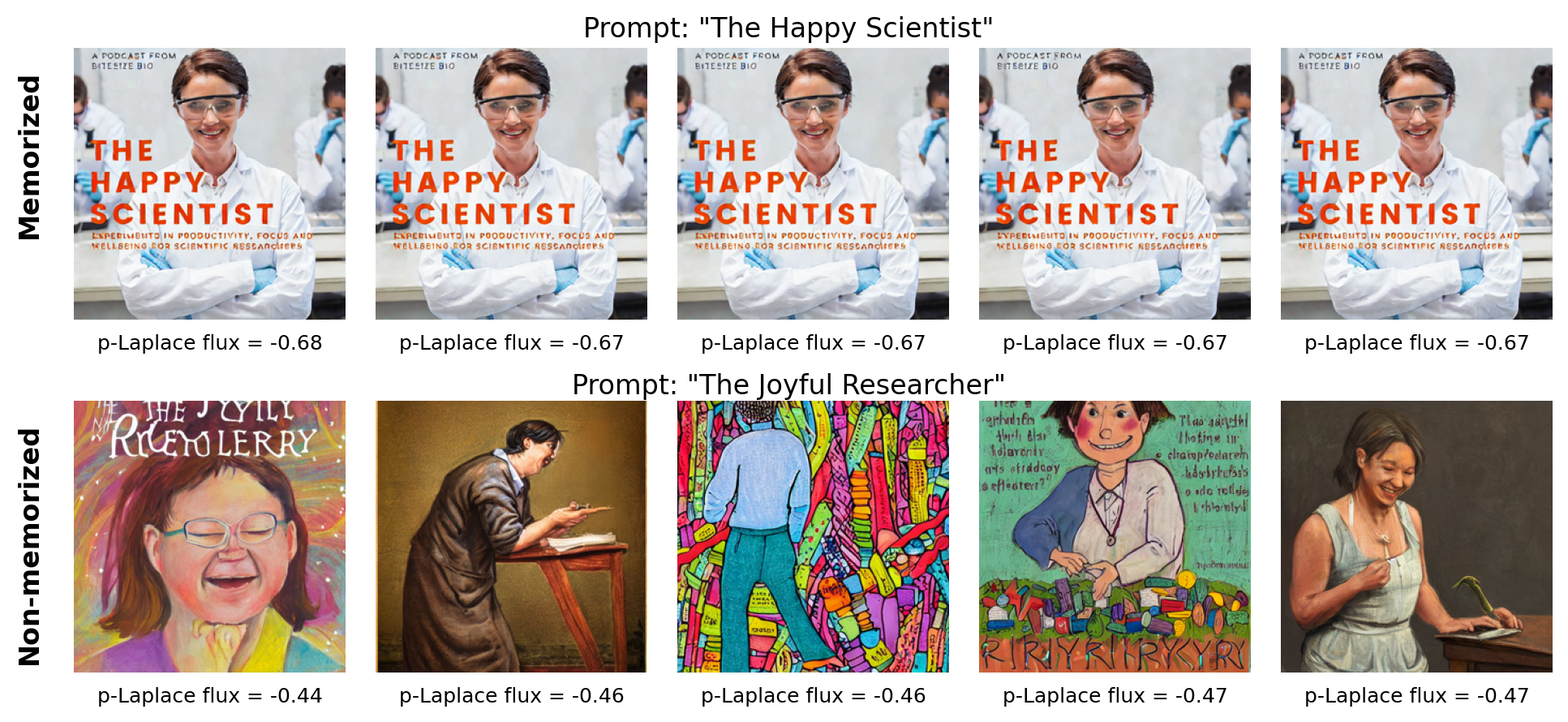}
\caption{Qualitative examples - two prompts with similar meanings where the first is memorized and the second is not, 5 generations per prompt. Their associated $1$-Laplace estimation reported beneath each generated image - showing clear distinction. See the Appendix, Fig. \ref{fig:additional_qualitative} for additional examples.} \label{fig:happy_scientist}
\end{figure*}

\begin{figure*}[htb]
\centering
\includegraphics[width=\textwidth]{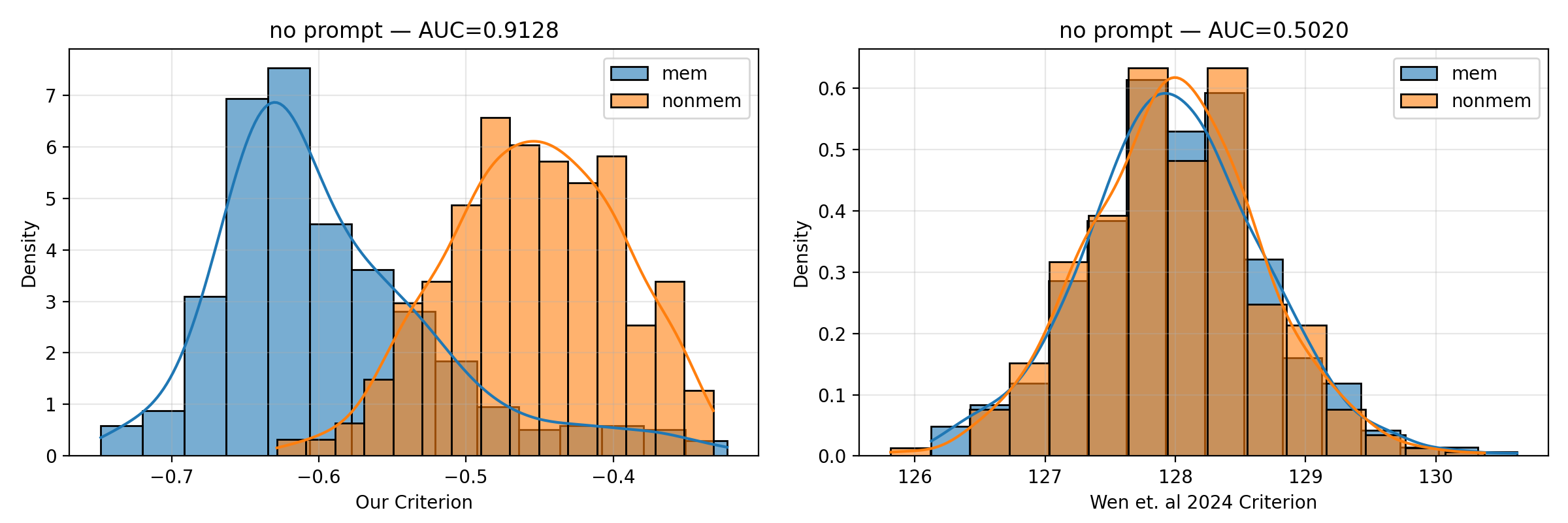}
\caption{\textcolor{black}{Testing our approach, using a pre-trained stable diffusion v1.4 to predict the score-function $\hat{s}$. $\hat{s}$ is plugged to the $1$-Laplace boundary integral formulation (Eq. \eqref{eq:boundary_formulation}) to distinguish memorized from non-memorized samples. The memorized prompts are taken from the dataset given by \cite{webster2023reproducible}, and the non-memorized prompts are LLM-generated prompts. Here we  compare results of our method vs the method proposed by \cite{wen2024detecting}, adapted to the promptless regime - where we do not have access to the memorized prompt. Comparison in both regimes (w/ and w/out prompt access) is reported in Table \ref{tab:auc_mem_nonmem}}.} \label{fig:comparison_histogram}
\end{figure*}

\subsection{Experiment 3: Error Bound Demonstration}
\textcolor{black}{
We now test the theoretical error bounds derived in Sec. \ref{subsec:error_bound} using the analytical GMM setting, which allows us to compute the approximation error explicitly. Specifically, we train a diffusion model on a 2-dimensional GMM, sample points from it, and compute $\abs{\overline{\Delta_ps}(x) - \overline{\Delta_p\hat{s}}(x)}$ for each sampled point $x$. This enables us to empirically validate the theoretical bound and to analyze its dependence on $p$ and the score approximation error $\delta$. Reults indeed demonstrate un upper-bound behaviour - see Fig. \ref{fig:error_bound_gmm}}

\subsection{Experiment 4: Large-Scale Image Generative Model Application}
\textcolor{black}{Here we apply our approach to a large-scale, pre-trained image diffusion model trained on LAION-5B. We choose Stable Diffusion v1.4 \cite{rombach2022high} as a testbed due to its popularity as well as well-studied memorization issues for certain prompts. We evaluate our method on a dataset of known memorized prompts (see \cite{webster2023reproducible}). For comparison, we also employ the method proposed by \cite{wen2024detecting}, which proposed using the difference of the scores between text-conditioned and unconditioned runs (often termed “Classifier-Free Guidance magnitude”). Concretely, let \(\hat{s}_\theta(x;c)\) denote the estimated score at step \(t\) when conditioned on prompt \(c\), and \(\hat{s}_\theta(x;\varnothing)\) the corresponding unconditional score. The per-step guidance magnitude is
\begin{equation}
\mathrm{Wen\,et.\, al, w/\text{-}prompt}\;:\;\bigl\|\hat{s}_\theta(x;c)\;-\;\hat{s}_\theta(x;\varnothing)\bigr\|_2.
\label{eq:cfg_mag_step}
\end{equation}
When the conditioning text \(c\) is unavailable (post-generation, promptless regime), we follow the same spirit and use the unconditional score magnitude as a proxy:
\begin{equation}
\mathrm{Wen\,et.\, al, No\text{-}prompt}\;:\;\bigl\|\hat{s}_\theta(x;\varnothing)\bigr\|_2,
\label{eq:no_prompt_mag}
\end{equation}
which is the minimal adaptation of this competitor that does not require access to \(c\). We compare these baselines to our \(1\)-Laplace boundary-flux estimator computed with and without $c$ - see Fig. \ref{fig:comparison_histogram} and table \ref{tab:auc_mem_nonmem}.
Following the conclusions of our experiments above, we employ the boundary integral formulation with $64$ spherical samples and \(p=1\)-Laplace. We use $T=500$ generation steps. We leave out the $\frac{|\partial B_R(x_0)|}{|B_R(x_0)|}$factor, since we are only interested in distinguishing memorized from non-memorized points, and this factor is uniform across samples (it only depends on the size of the neighborhood ball, which is fixed).}

\begin{table}[ht]
\caption{\textcolor{black}{AUC on memorized vs.\ non-memorized. Whole our method is comparable upon prompt access - it out-performs the competitor by 61\% AUC in the no-prompt regime}}\label{tab:auc_mem_nonmem}
\centering
{\normalsize
\begin{tabular}{@{}lll@{}}
\toprule
\textcolor{black}{Method} & \textcolor{black}{no prompt} & \textcolor{black}{w/ prompt} \\
\midrule
\textcolor{black}{Our Criterion}          & \textcolor{black}{0.913} & \textcolor{black}{0.958} \\
\textcolor{black}{Wen et al.\ Criterion}  & \textcolor{black}{0.502} & \textcolor{black}{0.957} \\
\botrule
\end{tabular}
}
\end{table}

\section{Conclusion}
We introduce a novel score-function analysis of diffusion models to approximate the $p$-Laplace operator of the unknown distribution of a sample dataset, and hypothesize that it can distinguish memorized from non-memorized generations. We propose two formulas to approximate the average of the $p$-Laplace over a neighborhood, approximating either a hyper-ball or a hyper-sphere integral. Our experiments in controlled synthetic settings indicate that the $1$-Laplace with the boundary integral formula is most effective. \textcolor{black}{We prove theoretical error bounds for this estimator, and verify it numerically in the same synthetic setting. Lastly, we demonstrate this approach's efficacy on stable-diffusion 1.4, a practical and widely used image generative model - on a dataset of 500 memorized and 500 non-memorized prompts}. To summarize, this work advances the understanding of diffusion models by offering new insights into the approximation of the \(p\)-Laplace operator and demonstrating its utility in characterizing memorization in image generative models, initiating a novel theoretical research direction for this timely task.

\backmatter

\begin{appendices}
\onecolumn

\section{\textcolor{black}{Proof of proposition 1}}\label{sec:err_bound_proof}
\textcolor{black}{
\noindent Fix $p \in[1, \infty], R>0$, and $x_0\in \mathbb{R}^d$.
We wish to bound the error of the $p$-Laplace average flux approximation on $\partial B_R(x_0)$. Namely, we wish to bound \begin{equation}
    \abs{\overline{\Delta_p s}(x_0) - \overline{\Delta_p \hat{s}}(x_0)}.
\end{equation}
Assume there exist $\delta>0$ and $0<m\leq M$ s.t. $\forall y \in \partial B_R(x_0)$, the following hold:
\begin{enumerate}[(a)]
    \item $\norm{s(y)-\hat{s}(y)}<\delta$ \label{assump:a}
    \item $m < \norm{s(y)}, \norm{\hat{s}(y)} \leq M$ \label{assump:b}
    \item $\forall t\in[0,1], \norm{t\cdot s(y)+(1-t)\cdot\hat{s}(y)} \geq m$ \label{assump:c}
\end{enumerate}
Then $\abs{\overline{\Delta_p s}(x_0) - \overline{\Delta_p \hat{s}}(x_0)} \leq \frac{\abs{\partial B_R(x_0)}}{\abs{B_R(x_0)}}C_p$, where
\begin{equation}
    C_p =
 \begin{dcases} 
      \delta M^{p-2}(p-1)  & p\geq 2 \\
      \delta m^{p-2}(3-p) & p<2 
   \end{dcases}
\end{equation}
}

\begin{proof}
\textcolor{black}{
By the triangle inequality, 
\begin{align*}
    \abs{\overline{\Delta_p s}(x_0) - \overline{\Delta_p \hat{s}}(x_0)} &= \frac{1}{\abs{B_R(x_0)}} \abs{\int_{\partial B_R(x_0)}(\abs{s(y)}^{p-2}s(y) - \abs{\hat{s}(y)}^{p-2}\hat{s}(y))\cdot n dy} \\ \leq
    & \frac{1}{\abs{B_R(x_0)}} \int_{\partial B_R(x_0)}\abs{\abs{s(y)}^{p-2}s(y) \cdot n - \abs{\hat{s}(y)}^{p-2}\hat{s}(y)\cdot n} dy
\end{align*}
We will begin by bounding the integrand.
Fix $y \in \partial B_R(x_0)$. Note that this fixes an outward unit normal $n$ to the sphere at $y$.
}

\textcolor{black}{
Denote $v:=s(y)$, $u:=\hat{s}(y)$, $k:=p-2$ for brevity.
By this notation, we want to bound $\abs{\norm{v}^kv\cdot n - \norm{u}^ku\cdot n}.$
Note that $u,v\in B_M(0)$ by ~\ref{assump:b}.
Define \[
F :B_M(0) \longrightarrow \mathbb{R} 
\]
\[
\quad x \longmapsto \norm{x}^kx\cdot n
\]
$F$ is smooth and defined on a convex set, hence we can apply the Mean Value Theorem (MVT). By MVT, we are guaranteed that $\exists c \in \left\{\, t u + (1-t)v \;\middle|\; t \in [0,1] \,\right\}$ such that \begin{equation}
    \abs{F(u)-F(v)} = \norm{\nabla F(c)}\cdot \norm{u-v}.
\end{equation}
By ~\ref{assump:a} we know that $\norm{u-v} < \delta$, so it remains to bound $\norm{\nabla F}$.}

\textcolor{black}{
One can verify that \begin{equation}
    \nabla F(x) = k\norm{x}^{k-2}(x\cdot n)x + \norm{x}^kn,
\end{equation}so \begin{align}
\norm{\nabla F(x)} 
  &= \norm{k\norm{x}^{k-2}(x\cdot n)x + \norm{x}^k n} \\
  &\le \norm{k\norm{x}^{k-2}(x\cdot n)x} + \norm{\norm{x}^k n} &&  \label{ineq:st1} \\
  &= \abs{k}\norm{x}^{k-2}\abs{x \cdot n}\norm{x} + \norm{x}^k\norm{n}  \\
  &\le \abs{k}\norm{x}^{k-2}\norm{x}\norm{n}\norm{x} + \norm{x}^k\norm{n}&& \label{ineq:st2} \\
  &= \abs{k}\norm{x}^k + \norm{x}^k = \norm{x}^k(1+\abs{k}) && \label{ineq:st3}
\end{align}
~\ref{ineq:st1} follows from the triangle inequality, ~\ref{ineq:st2} follows from the Cauchy-Schwartz inequality, and ~\ref{ineq:st3} follows from $\norm{n}=1.$}

\textcolor{black}{
We got that $\norm{x}^k(1+\abs{k})$ upper bounds $F$ on $B_M(0)$, and thus on $D:=\left\{\,x\in\mathbb{R}^d \;\middle|\; m \leq \norm{x} \leq M \, \right \} \subset B_M(0)$ which, by ~\ref{assump:b}, is our domain of interest.
From ~\ref{assump:c} we have $c \in D$, thus $\norm{\nabla F(c)} \leq \norm{c}^k(1+\abs{k}).$}

\textcolor{black}{
\paragraph{Case 1: $k \geq 0$}
If $k\geq 0$ then $\forall x\in D$, it holds that \begin{equation}
    \norm{\nabla F(x)} \leq \norm{x}^k(1+\abs{k}) \leq M^k(1+k)
\end{equation}
\paragraph{Case 2: $k < 0$}
If $k<0$ then $\forall x\in D$, it holds that \begin{equation}
    \norm{\nabla F(x)} \leq \norm{x}^k(1+\abs{k}) \leq m^k(1-k)
\end{equation}
In either case, we bounded $\norm{\nabla F(x)}$ on $D$. Specifically, we bounded $\norm{\nabla F(c)}$.}

\textcolor{black}{
Setting $k=p-2$, we get:\begin{equation}
    \abs{F(v)-F(u)} \leq \norm{\nabla F(c)}\cdot \delta\leq
    \begin{cases}
        \delta M^{p-2}(p-1), & \text{if } p>2, \\[6pt]
        \delta m^{p-2}(3-p), & \text{if } p<2.
    \end{cases}
\end{equation}
Namely,  $\abs{\abs{s(y)}^{p-2}s(y) \cdot n - \abs{\hat{s}(y)}^{p-2}\hat{s}(y)\cdot n} = \abs{F(v)-F(u)} \leq C_p$.
}

\textcolor{black}{
We conclude that $\forall y\in B_R(x_0)$, $\abs{\abs{s(y)}^{p-2}s(y)\cdot n - \abs{\hat{s}(y)}^{p-2}\hat{s}(y)\cdot n} \leq C_p$.}

\textcolor{black}{
Thus, we obtain \begin{align*}
     \abs{\overline{\Delta_p s}(x_0) - \overline{\Delta_p \hat{s}}(x_0)} \leq
     \frac{1}{\abs{B_R(x_0)}} \int_{\partial B_R(x_0)}\abs{\abs{s(y)}^{p-2}s(y) \cdot n - \abs{\hat{s}(y)}^{p-2}\hat{s}(y)\cdot n} dy &\leq \\ \frac{1}{\abs{B_R(x_0)}} \int_{\partial B_R(x_0)}C_p = \frac{\abs{\partial B_R(x_0)}}{\abs{B_R(x_0)}}C_p
\end{align*}}

\end{proof}

\section{\textcolor{black}{Additional Qualitative Examples}}
\begin{figure*}[hb]
\centering
\includegraphics[width=1\textwidth]{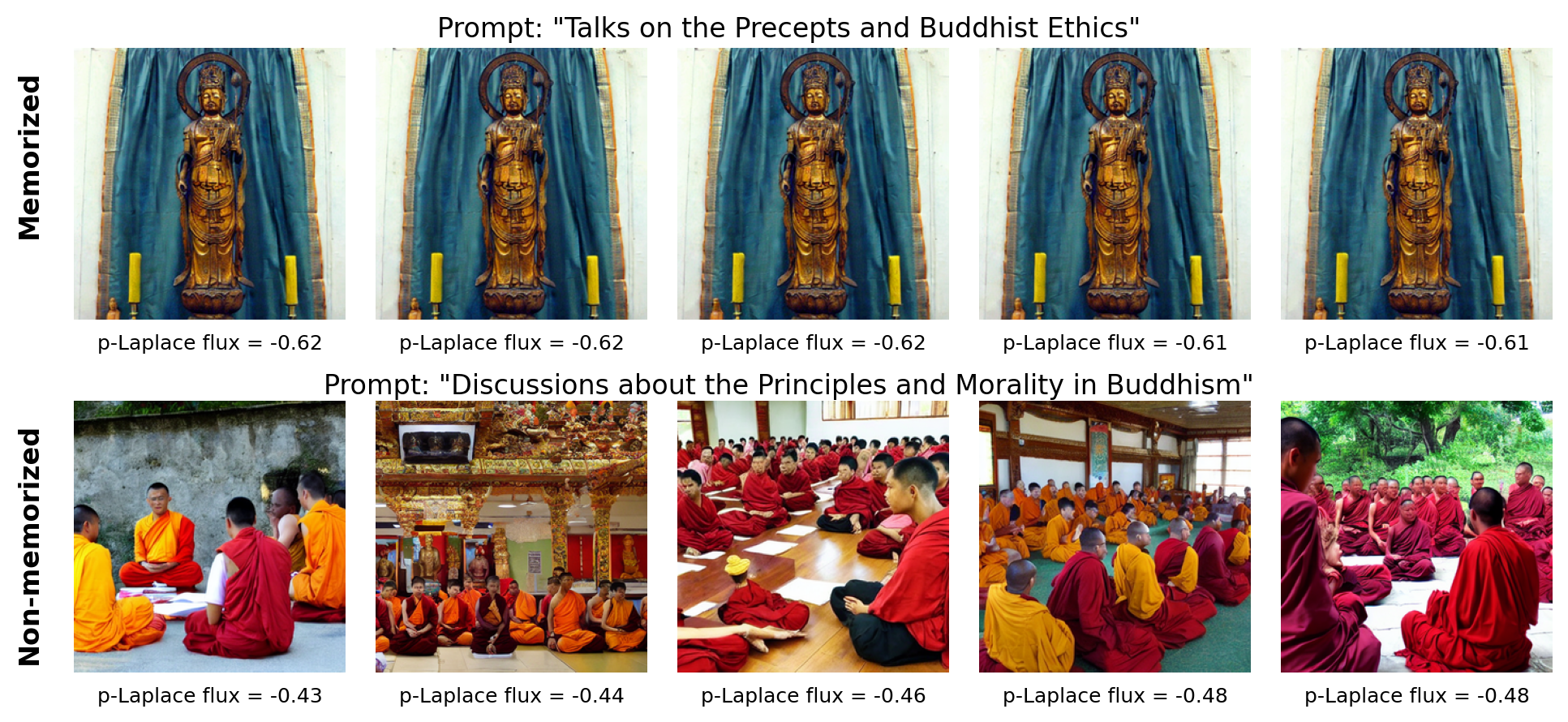}
\includegraphics[width=1\textwidth]{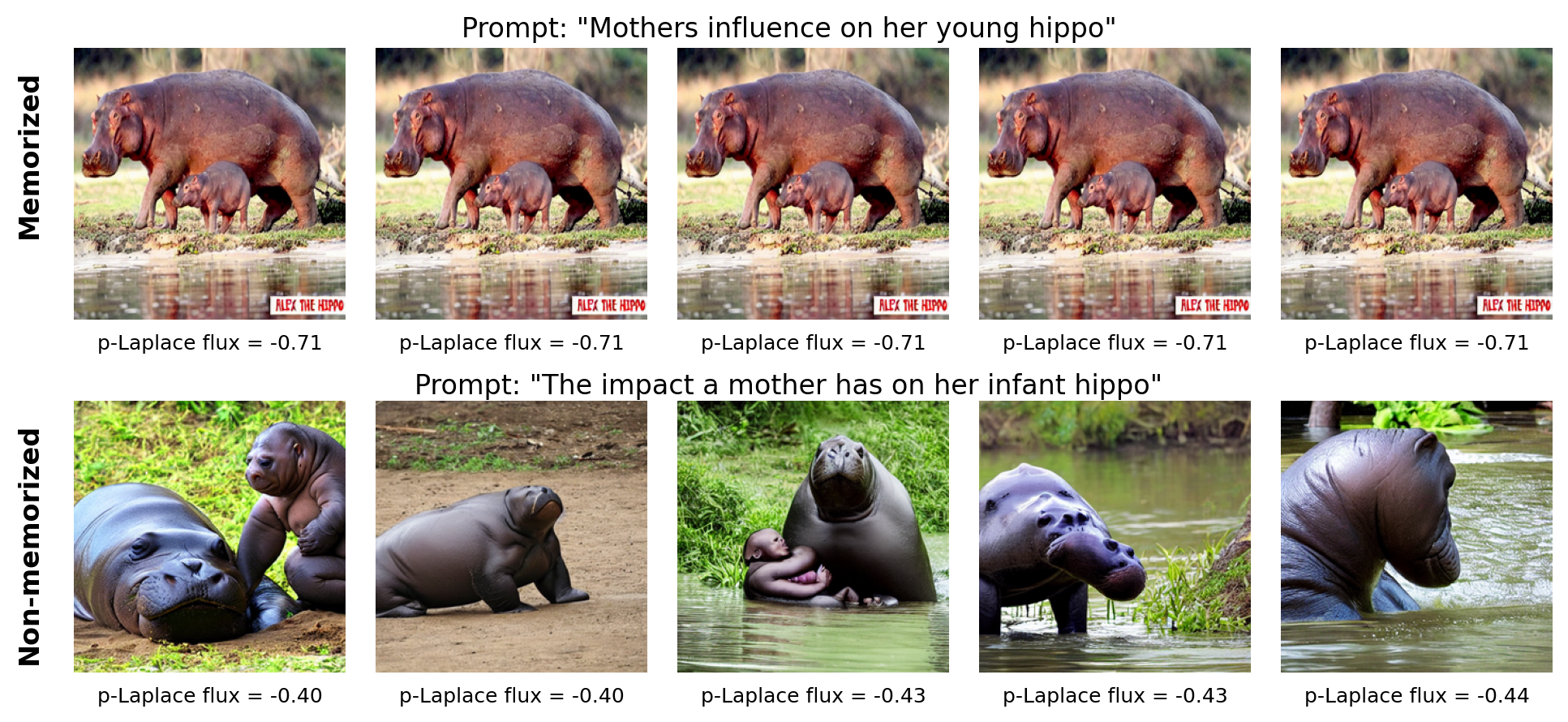}
\includegraphics[width=1\textwidth]{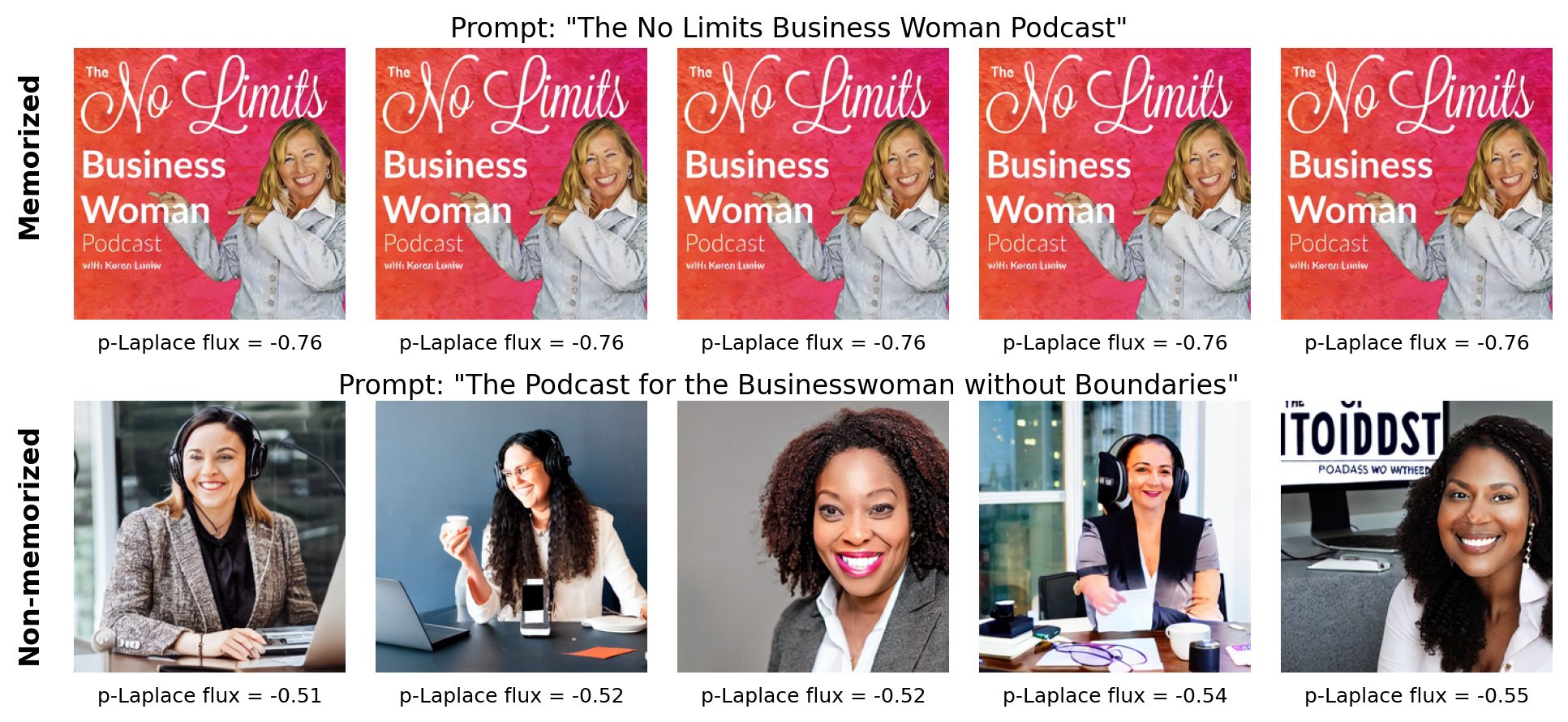}
\caption{\textcolor{black}{Additional qualitative examples - similarly to Fig. \ref{fig:happy_scientist}}} \label{fig:additional_qualitative}
\end{figure*}

\textcolor{black}{Fig. \ref{fig:additional_qualitative} provides further examples of memorized vs nonmemroized generations and their associated $\Delta_1$e.}

\end{appendices}


\end{document}